\documentclass{article}

\usepackage{latexsym,amsmath,amssymb}

\makeatletter

 \twocolumn
\def\addcontentsline#1#2#3{}

\long\def\icmltitle#1{%
  {\center\baselineskip 18pt
                       \toptitlebar{\Large\bf #1}\bottomtitlebar}
}
\def\toptitlebar{\hrule height1pt \vskip .25in}
\def\bottomtitlebar{\vskip .22in \hrule height1pt \vskip .3in}
\def\icmlauthor#1#2{\par {\bf #1} \hfill {\sc #2}}
\long\def\icmladdress#1{\par\vskip 0.03in #1 \vskip 0.10in}

\renewenvironment{abstract}
   {%
\centerline{\large\bf Abstract}
    \vspace{-0.12in}\begin{quote}}
   {\par\end{quote}\vskip 0.12in}

\def\@startsection#1#2#3#4#5#6{\if@noskipsec \leavevmode \fi
   \par \@tempskipa #4\relax
   \@afterindenttrue
   \ifdim \@tempskipa <\z@ \@tempskipa -\@tempskipa \fi
   \if@nobreak \everypar{}\else
     \addpenalty{\@secpenalty}\addvspace{\@tempskipa}\fi \@ifstar
     {\@ssect{#3}{#4}{#5}{#6}}{\@dblarg{\@sict{#1}{#2}{#3}{#4}{#5}{#6}}}}

\def\@sict#1#2#3#4#5#6[#7]#8{\ifnum #2>\c@secnumdepth
     \def\@svsec{}\else
     \refstepcounter{#1}\edef\@svsec{\csname the#1\endcsname}\fi
     \@tempskipa #5\relax
      \ifdim \@tempskipa>\z@
        \begingroup #6\relax
          \@hangfrom{\hskip #3\relax\@svsec.~}{\interlinepenalty \@M #8\par}
        \endgroup
       \csname #1mark\endcsname{#7}\addcontentsline
         {toc}{#1}{\ifnum #2>\c@secnumdepth \else
                      \protect\numberline{\csname the#1\endcsname}\fi
                    #7}\else
        \def\@svsechd{#6\hskip #3\@svsec #8\csname #1mark\endcsname
                      {#7}\addcontentsline
                           {toc}{#1}{\ifnum #2>\c@secnumdepth \else
                             \protect\numberline{\csname the#1\endcsname}\fi
                       #7}}\fi
     \@xsect{#5}}

\def\@sect#1#2#3#4#5#6[#7]#8{\ifnum #2>\c@secnumdepth
     \def\@svsec{}\else
     \refstepcounter{#1}\edef\@svsec{\csname the#1\endcsname\hskip 0.4em }\fi
     \@tempskipa #5\relax
      \ifdim \@tempskipa>\z@
        \begingroup #6\relax
          \@hangfrom{\hskip #3\relax\@svsec}{\interlinepenalty \@M #8\par}
        \endgroup
       \csname #1mark\endcsname{#7}\addcontentsline
         {toc}{#1}{\ifnum #2>\c@secnumdepth \else
                      \protect\numberline{\csname the#1\endcsname}\fi
                    #7}\else
        \def\@svsechd{#6\hskip #3\@svsec #8\csname #1mark\endcsname
                      {#7}\addcontentsline
                           {toc}{#1}{\ifnum #2>\c@secnumdepth \else
                             \protect\numberline{\csname the#1\endcsname}\fi
                       #7}}\fi
     \@xsect{#5}}

\def\section{\@startsection{section}{1}{\z@}{-0.12in}{0.02in}
             {\large\bf\raggedright}}
\def\subsection{\@startsection{subsection}{2}{\z@}{-0.10in}{0.01in}
                {\normalsize\bf\raggedright}}
\def\subsubsection{\@startsection{subsubsection}{3}{\z@}{-0.08in}{0.01in}
                {\normalsize\sc\raggedright}}
\def\paragraph{\@startsection{paragraph}{4}{\z@}{1.5ex plus
  0.5ex minus .2ex}{-1em}{\normalsize\bf}}
\def\subparagraph{\@startsection{subparagraph}{5}{\z@}{1.5ex plus
  0.5ex minus .2ex}{-1em}{\normalsize\bf}}

\skip\footins 9pt
\def\footnoterule{\kern-3pt \hrule width 0.8in \kern 2.6pt }
\labelwidth\leftmargini\advance\labelwidth-\labelsep \labelsep 5pt

\def\@listi{\leftmargin\leftmargini}
\def\@listii{\leftmargin\leftmarginii
   \labelwidth\leftmarginii\advance\labelwidth-\labelsep
   \topsep 2pt plus 1pt minus 0.5pt
   \parsep 1pt plus 0.5pt minus 0.5pt
   \itemsep \parsep}
\def\@listiii{\leftmargin\leftmarginiii
    \labelwidth\leftmarginiii\advance\labelwidth-\labelsep
    \topsep 1pt plus 0.5pt minus 0.5pt
    \parsep \z@ \partopsep 0.5pt plus 0pt minus 0.5pt
    \itemsep \topsep}
\def\@listiv{\leftmargin\leftmarginiv
     \labelwidth\leftmarginiv\advance\labelwidth-\labelsep}
\def\@listv{\leftmargin\leftmarginv
     \labelwidth\leftmarginv\advance\labelwidth-\labelsep}
\def\@listvi{\leftmargin\leftmarginvi
     \labelwidth\leftmarginvi\advance\labelwidth-\labelsep}

\belowdisplayskip \abovedisplayskip
\def\@normalsize{\@setsize\normalsize{11pt}\xpt\@xpt}
\def\small{\@setsize\small{10pt}\ixpt\@ixpt}
\def\footnotesize{\@setsize\footnotesize{10pt}\ixpt\@ixpt}
\def\scriptsize{\@setsize\scriptsize{8pt}\viipt\@viipt}
\def\tiny{\@setsize\tiny{7pt}\vipt\@vipt}
\def\large{\@setsize\large{14pt}\xiipt\@xiipt}
\def\Large{\@setsize\Large{16pt}\xivpt\@xivpt}
\def\LARGE{\@setsize\LARGE{20pt}\xviipt\@xviipt}
\def\huge{\@setsize\huge{23pt}\xxpt\@xxpt}
\def\Huge{\@setsize\Huge{28pt}\xxvpt\@xxvpt}

\newsavebox\captionbox\newdimen\captionboxwid

\long\def\@makecaption#1#2{
 \vskip 10pt
        \baselineskip 11pt
        \setbox\@tempboxa\hbox{#1. #2}
        \ifdim \wd\@tempboxa >\hsize
        \sbox{\captionbox}{\small\sl #1.~}
        \captionboxwid=\wd\captionbox
        \usebox\captionbox {\footnotesize #2}
        \else
          \centerline{{\small\sl #1.} {\small #2}}
        \fi}

\def\fnum@figure{Figure \thefigure}
\def\fnum@table{Table \thetable}

\def\texitem#1{\par\noindent\hangindent 12pt
               \hbox to 12pt {\hss #1 ~}\ignorespaces}

\long\def\comment#1{}

\makeatother

\def\emcite{\cite}\def\aunpcite{\cite}\def\npcite{\cite}\def\yrcite{\cite}
\newtheorem{Def1}{Definition}
\newtheorem{Lemma}[Def1]{Lemma}

\newtheorem{Theorem}[Def1]{Theorem}
\newtheorem{Cor}[Def1]{Corollary}

\newtheorem{Ex1}[Def1]{Example}

\def\defaultskip{}

\newenvironment{Proof}[1][.]{\defaultskip\noindent \textbf{Proof}#1 }
  {\hspace*{0mm}\hfill $\Box$ \defaultskip}

\def\beq{\begin{equation}}
\def\eeq{\end{equation}}
\def\beqn{\begin{displaymath}}
\def\eeqn{\end{displaymath}}
\def\bqa{\begin{eqnarray}}
\def\eqa{\end{eqnarray}}
\def\bqan{\begin{eqnarray*}}
\def\eqan{\end{eqnarray*}}

\def\calA{\mathcal A}
\def\calB{\mathcal B}

\def\NNN{\mathbb N}
\def\RRR{\mathbb R}

\def\Expect{{\mathbf E}}
\def\Prob{{\mathbf P}}

\def\leqt{_{1:t}}
\def\ltt{_{<t}}

\def\_norm{_\mathrm{norm}}

\newcommand{\zwidths}[1]{\rlap{$\scriptstyle #1$}}

\def\for_all{\mbox{ for all }}
\def\such_that{\mbox{ such that }}

\def\und{\mbox{ and }}

\def\xwidehat{}

\def\FOE{{\xwidehat{\mbox{\textit{F\kern-0.12emo\kern-0.08emE}}}}}
\def\FPL{{\xwidehat{\mbox{\textit{F\kern-0.08emP\kern-0.08emL}}}}}
\def\IFPL{{\xwidehat{\mbox{\textit{IF\kern-0.08emP\kern-0.08emL}}}}}
\def\best{{\mbox{\scriptsize\textit{b\kern-0.08eme\kern-0.05ems\kern-0.05emt}}}}
\def\foe{^{\xwidehat{\mbox{\scriptsize\textit{F\kern-0.13emo\kern-0.13emE}}}}}
\def\foett{^{\xwidehat{\mbox{\scriptsize\textit{F\kern-0.13emo\kern-0.13emE}}}_{\smash{\tilde T}}}}
\def\fpl{^{\xwidehat{\mbox{\scriptsize\textit{F\kern-0.15emP\kern-0.15emL}}}}}
\def\ifpl{^{\xwidehat{\mbox{\scriptsize\textit{I\kern-0.08emF\kern-0.15emP\kern-0.15emL}}}}}
\def\foetau{^{\xwidehat{\mbox{\scriptsize\textit{F\kern-0.13emo\kern-0.13emE}}}{}^\tau}}
\def\foetttau{^{\xwidehat{\mbox{\scriptsize\textit{F\kern-0.13emo\kern-0.13emE}}}{}_{\smash{\tilde T}}^\tau}}
\def\fpltau{^{\xwidehat{\mbox{\scriptsize\textit{F\kern-0.15emP\kern-0.15emL}}}{\!}^\tau}}
\def\ifpltau{^{\xwidehat{\mbox{\scriptsize\textit{I\kern-0.08emF\kern-0.15emP\kern-0.15emL}}}{\!}^\tau}}
\def\sfoe{^{\xwidehat{\mbox{\tiny\textit{F\kern-0.12emo\kern-0.18emE}}}}}
\def\sfpl{^{\xwidehat{\mbox{\tiny\textit{F\kern-0.2emP\kern-0.2emL}}}}}
\def\sifpl{^{\xwidehat{\mbox{\tiny\textit{I\kern-0.1emF\kern-0.2emP\kern-0.2emL}}}}}

\def\eins{1\hspace{-0.23em}{\rm I}}
\def\leqT{_{1:T}}
\def\ltT{_{<T}}
\def\tfrac#1#2{{\textstyle\frac{#1}{#2}}}
\def\tsqrt#1{{\textstyle\sqrt{#1}}}
\def\e{{\rm e}}
\def\approxleq{\;\mbox{\raisebox{-0.8ex}{$\stackrel{\displaystyle<}{\sim}$}}\;}
\def\wprob{\mbox{ w.p. }}

\def\approxleq{\mbox{\raisebox{-0.7ex}{$\stackrel{\displaystyle<}{\scriptstyle\sim}$}}}
\def\ii{\hspace*{1em}}

\newlength\algowidth
\def\,{\mskip 3mu} \def\>{\mskip 4mu plus 2mu minus 4mu} \def\;{\mskip 5mu plus 5mu} \def\!{\mskip-3mu}
\def\dispmuskip{\thinmuskip= 3mu plus 0mu minus 2mu \medmuskip=  4mu plus 2mu minus 2mu \thickmuskip=5mu plus 5mu minus 2mu}
\def\textmuskip{\thinmuskip= 0mu                    \medmuskip=  1mu plus 1mu minus 1mu \thickmuskip=2mu plus 3mu minus 1mu}
\textmuskip
\def\beq{\dispmuskip\begin{equation}}    \def\eeq{\end{equation}\textmuskip}
\def\beqn{\dispmuskip\begin{displaymath}}\def\eeqn{\end{displaymath}\textmuskip}
\def\bqa{\dispmuskip\begin{eqnarray}}    \def\eqa{\end{eqnarray}\textmuskip}
\def\bqan{\dispmuskip\begin{eqnarray*}}  \def\eqan{\end{eqnarray*}\textmuskip}
\def\baq#1\eaq{\dispmuskip\begin{align}#1\end{align}\textmuskip}
\def\baqn#1\eaqn{\dispmuskip\begin{align*}#1\end{align*}\textmuskip}

\def\Loss{\ell}
\def\loss{\ell}
\def\LOSS{L\!}
\def\LOSShat{\hat{L}\!}

\begin{document}

\twocolumn[IDSIA-01-05 \hfill 16 January 2005
\icmltitle{Master Algorithms for Active Experts Problems \\
           based on Increasing Loss Values}
\icmlauthor{Jan Poland}{jan@idsia.ch}
\icmlauthor{Marcus Hutter}{marcus@idsia.ch}
\icmladdress{IDSIA, Galleria 2, CH-6928 Manno-Lugano,
Switzerland \hfill {\sc www.idsia.ch}}
\vskip 0.3in
]

\begin{abstract}
We specify an experts algorithm with the following
characteristics: (a) it uses only feedback from the actions
actually chosen (bandit setup), (b) it can be applied with
countably infinite expert classes, and (c) it copes with
losses that \emph{may grow in time} appropriately slowly. We
prove loss bounds against an adaptive adversary. From this, we
obtain master algorithms for ``active experts problems", which
means that the master's actions may influence the behavior of
the adversary. Our algorithm can significantly outperform
standard experts algorithms on such problems. Finally, we
combine it with a universal expert class. This results in a
(computationally infeasible) \emph{universal master algorithm}
which performs -- in a certain sense -- almost as well as any
computable strategy, for any online problem.

\end{abstract}

\section{Introduction}

Expert algorithms have been popular since about fifteen years
ago \cite{Littlestone:89}. They are appropriate for online
prediction or repeated decision making or repeated game
playing (we call these setups
\emph{online problems} for brevity), based on a class of
``experts". In each round, each expert gives a recommendation.
From this, we derive a master decision. After that, losses (or
rewards) are assigned to each expert by the environment, also
called adversary. Our goal is to perform almost as well as the
best expert in hindsight in the long run. In other words, we
try to minimize the regret.

The early papers deal with the full information game, where we
get to know the losses of each expert after each round. The
analysis holds for the \emph{worst case}, where the
environment is fully adversarial and tries to maximize our
regret in the long run. Later, \emcite{Auer:95} gave a
worst-case analysis for the \emph{bandit} setup, where the
master algorithm knows only the loss of its own decision after
each round. This has been further generalized to
\emph{label-efficient prediction} \cite{Helmbold:97} and
\emph{partial monitoring} \cite{Cesa:04partial}.

Recently, \emcite{Farias:03} introduced a \emph{strategic
experts algorithm} which performs well for a broader class of
environments. The algorithm has still asymptotically optimal
properties against a worst-case adversary. Additionally, it
may perform much better than a standard experts algorithm in
more favorable situations, when the actions influence the
behavior of the environment. We refer to these as
\emph{active experts problems}. One
example is the repeated prisoner's dilemma when the opponent
is willing to cooperate under certain conditions (see Section
\ref{sec:active} for some details). However,
\aunpcite{Farias:03} give only asymptotic guarantees, but no
convergence rate.

In this paper, we introduce a different algorithm for active
experts problems with the same asymptotic guarantees, but in
addition a convergence rate (of $t^{-\frac{1}{10}}$) is shown.
Both algorithm and analysis are assembled from a standard
``toolkit", basing on \emcite{Kalai:03,McMahan:04}. The basic
idea is the following: We use the bandit experts algorithm by
\aunpcite{McMahan:04}, but allow the losses to increase with
time $t$. This allows us to give control to one expert for an
\emph{increasing} period of time steps.

Secondly, we generalize our analysis to the case of
\emph{infinitely many} experts, basing on \emcite{Hutter:04expert}.
The master algorithm stays computable (if the experts are),
since only a finite (with time increasing) number of experts
is involved. Allowing infinitely many experts
also permits to define a
\emph{universal expert class} by means of all programs on some
universal Turing machine. (This construction is quite common
in Algorithmic Information Theory, see e.g.\
\npcite{Hutter:04uaibook}.) Thus, we obtain a \emph{universal
master algorithm}, which we show to perform in a certain sense
almost as well as \emph{any computable} strategy on \emph{any
online problem}. Thus, we introduce a new approach to
universal artificial intelligence, which is in a sense dual to
the AIXI model based on Bayesian learning
\cite{Hutter:04uaibook}. Although the master algorithm is
computable, the resulting universal agent is not (like the
AIXI model), since the experts may be non-responsive.

The paper is structured as follows. Section
\ref{sec:algorithm} introduces the problem setup, the
notation, and the algorithm. In Sections \ref{sec:uniform} and
\ref{sec:arbitrary}, we give the (worst-case)
analysis for finite and infinite expert classes. The
implications to active experts problems and a universal master
algorithms are given in Section \ref{sec:active}. Section
\ref{sec:discussion} contains discussion and conclusions.

\section{The Algorithm}
\label{sec:algorithm}

Our task is an online decision problem. That is, we have to
make a sequence of decisions, each of which results in a
certain loss we incur. ``We" is an abbreviation for the master
algorithm which is to be designed. For concreteness, you may
imagine the task of playing a game repeatedly. In each round,
i.e.\ at each time step $t$, we have access to the
recommendations of
$n\in\NNN\cup\{\infty\}$ ``experts" or strategies. We do not
specify what exactly a ``recommendation" is -- we just follow
the advice of one expert. \emph{Before} we reveal our move,
the adversary has to assign losses
$\loss_t^i\geq 0$ to
\emph{all} experts $i$. There is an upper bound
$B_t$ on the maximum loss the adversary
may use, i.e.\ $\ell_t\in[0,B_t]^n$. This quantity may depend
on $t$ and is known to us. After the move, only the loss of
the selected expert
$i$ is revealed. This is the \emph{bandit setup}, as opposed
to the full information game where we get to know the losses
\emph{all} experts.
Our goal is to perform nearly as well as the best available
strategy in terms of cumulative loss, after any number $T$ of
time steps which is not known in advance. The difference
between our loss and the loss of some expert is also termed
\emph{regret}. We consider the general case of an
\emph{adaptive} adversary, which may assign losses depending
on our past decisions.

If there is a finite number $n$ of experts or strategies, then
it is common to give no prior preferences to any of them.
Formally, we define \emph{prior weights}
$w^i=\frac{1}{n}$. Moreover, we define the complexity of
expert $i$ as $k^i=-\ln w^i$. This arises in the full
observation game, where the regret can be bounded by some
function of the best expert's complexity. On the other hand,
if there are reasons not to trust all strategies equally in
the beginning, we may use a non-uniform prior $w$. This is
mandatory for infinitely many experts. We then require $w^i>0$
for all experts $i$ and $\sum_i w^i\leq 1$.

Our algorithm ``Follow or Explore" ($\FOE$) builds on McMahan
and Blum's online geometric optimization algorithm. (For
finite $n$ and uniform prior, it even is their algorithm, save
for the adaptive parameters.) It is a bandit version of a
``Follow the Perturbed Leader" experts algorithm. This
approach to online prediction and playing repeated games has
been pioneered by \emcite{Hannan:57}. For the full observation
game, \emcite{Kalai:03} gave a very elegant analysis which is
distinct from the standard analysis of exponential weighting
schemes. It is particularly handy if the learning rate is
dynamic rather than fixed in advance. A dynamic learning rate
is necessary if there is no target time
$T$ known in advance.

\begin{figure}[t!]
\begin{center}
\fbox{
\begin{minipage}{\algowidth}
For $t=1,2,3,\ldots$\\
Sample $r_t\in\{0,1\}$ independently s.t.
$P[r_t=1]=\gamma_t$\\
If $r_t=0$ Then\\
\ii Play $\FPL(t)$'s decision ($I_t\foe:=I_t\fpl$)\\
\ii Set $\hat\loss_t^i=0$ for all $1\leq i\leq n$\\
Else\\
\ii Sample $I_t\foe\!\in\!\{1...n\}$ uniformly \& play $I\!:=\!I_t\foe$\\
\ii Let $\hat\loss_t^I=\loss_t^I n/\gamma_t$ and $\hat\loss_t^i=0$ for all $i\neq I$
\end{minipage}}
\caption{The algorithm $\FOE$}
\label{fig:foe}
\end{center}
\end{figure}

The algorithm is composed of two standard ingredients:
\emph{exploration} and \emph{follow the (perturbed) leader}.
Since we are playing the bandit game (as opposed to the full
information game), we need to explore sufficiently. Otherwise,
there could be a strategy which we think is poor (and thus
never play), but in reality it is good. At each time step $t$,
we decide randomly according to some exploration rate
$\gamma_t\in(0,1)$ whether to explore or not. If so, we choose an
expert according to the uniform distribution (or the prior
distribution, compare (\ref{eq:unonu}), in case of non-uniform
priors). After observing the loss of the selected expert, we
want to give an
\emph{unbiased estimate} of the true loss vector. We achieve
that by dividing the observed loss by the probability of
exploring this expert, and estimate the unobserved losses of
all other experts by zero. We call the resulting loss vector
$\hat\loss_t$.

\begin{figure}[t!]
\begin{center}
\fbox{
\begin{minipage}{\algowidth}
Sample $q_t^i\stackrel{d.}{\sim}\mbox{\textit{Exp}}$
independently for $1\leq i\leq n$\\
select and play $I_t\fpl=\arg\min\limits_{1\leq i\leq
n}\{\eta_t\hat \Loss^i\ltt+k^i-q_t^i\}$
\end{minipage}}
\caption{The algorithm $\FPL(t)$}
\label{fig:fpl}
\end{center}
\end{figure}

When not exploring, we follow some strategy which performed
well in the past. It may be not advisable to pick always the
\emph{best} strategy so far - the adversary could fool us in
this case. Instead we introduce a \emph{perturbation} for each
expert and follow the advice of the strategy with the best
perturbed score. In order to assign a score to each expert,
note that we have only access to the \emph{estimated} losses
$\hat\loss_t$. Let $\hat \Loss\ltT^i=\sum_{t=1}^{T-1} \hat\loss_t^i$
be the estimated cumulative past loss of expert $i$. Then his
complexity-penalized score is defined as $\eta_T \hat
\Loss\ltT^i+k^i$, i.e.\ high scores are bad.
Here, $\eta_T>0$ is the \emph{learning rate}. The perturbed
score is then given by $\eta_T \hat \Loss\ltT^i+k^i-q^i$,
where the perturbations $q^i$ are chosen independently
exponentially distributed. This ensures a convenient analysis.

The algorithms ``Follow or Explore" $\FOE$ and ``Follow the
perturbed Leader" $\FPL$ are fully specified in Figures
\ref{fig:foe} and \ref{fig:fpl}. Note that each time
randomness is used, it is assumed to be \emph{independent} of
the past randomness. Note also that all algorithms occurring
in this paper work with the \emph{estimated losses}
$\hat\loss$. We may evaluate their performance in terms of true or
estimated losses, this is specified in the notation. E.g.\ for
the true loss of $\FPL$ up to and including time $T$ we write
$\LOSS\fpl=\Loss\fpl\leqT$, while the estimated loss is
$\LOSShat\fpl=\hat\Loss\fpl\leqT$.

\section{Analysis for Uniform Prior}
\label{sec:uniform}

In this section we assume a uniform prior $w\equiv\frac{1}{n}$
over finitely many experts. (The general case is treated in
the next section.) We assume that
$B_t\geq 0$ is some sequence of upper bounds on the true
losses,
$\gamma_t\in(0,1)$ is a sequence of exploration rates,
and $\eta_t>0$ is a \emph{decreasing} sequence of learning
rates.

The analysis is according to the following diagram:
\beq
\label{eq:diagram}
\LOSS\foe \approxleq \Expect \LOSS\foe \approxleq \Expect
\LOSS\fpl
\approxleq \Expect \LOSShat\fpl
\approxleq \Expect \LOSShat\ifpl
\approxleq \LOSShat^{\xwidehat\best} \approxleq \LOSS^{\best}
\!
\eeq
The symbol $\LOSS\,$ is used informally for the cumulative
loss $\ell\leqT$. Each ``$\approxleq$" means that we bound the
quantity on the left by the quantity on the right plus some
additive terms. The first and the last expressions are the
losses of the
$\FOE$ algorithm and the best expert, respectively. The
intermediate quantities belong to different algorithms, namely
$\FOE$, $\FPL$, and a third one called $\IFPL$ for
``infeasible" FPL \cite{Kalai:03}. $\IFPL$ is the same as
$\FPL$ except that it has access to an oracle providing the
current estimated loss vector $\hat\loss_t$ (hence
infeasible). Then it assigns scores of $\eta_t\hat
\Loss^i\leqt+k^i-q_t^i$ instead of
$\eta_t\hat \Loss^i\ltt+k^i-q_t^i$. We assume that $\IFPL$ uses
the same randomization as $\FPL$ (i.e.\ the respective $q_t$
are the same).

The randomization of $\FOE$ and $\FPL$ gives rise to two
filters of $\sigma$-algebras. By $\calA_t$ for $t\geq 0$ we
denote the $\sigma$-algebra generated by the $\FOE$'s
randomness
$\{u_{1:t},r_{1:t}\}$ up to time $t$. We may also write
$\calA=\bigcup_{t\geq 0}\calA_t$. Similarly,
$\calB_t$ is the $\sigma$-algebra generated
by the $\FOE$'s \emph{and} $\FPL$'s randomness up to time $t$
(i.e.\ $\calB_t\widehat=\{u_{1:t},r_{1:t},q_{1:t}\}$).
Then clearly $\calA_t\subset\calB_t$ for each $t$.

The arguments below rely on \emph{conditional expectations} --
the expectations in (\ref{eq:diagram}) should also be
understood conditional. In particular we will often need the
conditional expectations with respect to $\FOE$'s past
randomness
$\calA_{t-1}$, abbreviated as
\beqn
  \Expect_t[X]:=\Expect[X|\calA_{t-1}],
\eeqn
where $X$ is some random variable. Then $\Expect_t[X]$ is an
$\calA_{t-1}$-measurable random variable, meaning that its value
is determined for fixed past randomness $\calA_{t-1}$. Note in
particular that the estimated loss vectors $\hat \loss_t^i$
are random vectors which depend on $\FOE$'s randomness
$\calA_t$ up to time $t$ (only). In this way, $\FOE$'s (and
$\FPL$'s and
$\IFPL$'s) actions depend on $\FOE$'s past randomness. Note,
however, that they do not depend on $\FPL$'s randomness
$q_{1:t}$. Finally, $I_t\foe$ and $\loss_t\foe$ are $\calA'_t$
measurable, i.e.\ depend on $u_{<t},r_{<t},q_t$, but are
independent of $q_{<t}$.

We now start proving the diagram (\ref{eq:diagram}). It is
helpful to consider each intermediate algorithm as a
stand-alone procedure which is actually executed (with an
oracle if necessary) and has the asserted performance
guarantees (e.g.\ in terms of expected losses).

\begin{Lemma} \label{lemma:foefoe}
$\big[\LOSS\foe \approxleq \Expect \LOSS\foe\big]$ For
each $T\geq 1$ and $\delta_T\in(0,1)$, with probability at
least $1-\tfrac{\delta_T}{2}$, we have
\baqn
\Loss\foe\leqT\leq\sum_{t=1}^t\Expect_t\loss\foe_t
+\tsqrt{(2\ln\tfrac{4}{\delta_T}){\textstyle\sum\nolimits_{t=1}^{T}
B_t^2}}.
\eaqn
\end{Lemma}

\begin{Proof} The sequence of random variables
$X_T=\sum_{t=1}^T\big[\loss\foe_t-
\Expect_t\loss\foe_t\big]$ is a martingale with
respect to the filter $\calB_t$ (not $\calA_t$!). In order to
see this, observe
$\Expect[\loss\foe_T|\calB_{T-1}]=
\Expect\big(\Expect[\loss\foe_T|\calA_{T-1}]\big|\calB_{T-1}\big)$
and
$\Expect[\loss\foe_t|\calB_{T-1}]=\loss_t\foe$
for $t<T$, which implies
\baqn
\Expect&(X_T|\calB_{T-1}) = \\
&=\sum\nolimits_{t=1}^T\left(
\Expect[\loss\foe_t|\calB_{T-1}]-
\Expect\big[\Expect[\loss\foe_t|\calA_{t-1}]\big|\calB_{T-1}\big]\right)
\\ &=
\sum_{t=1}^{T-1}\nolimits\left(
\loss\foe_t-
\Expect[\loss\foe_t|\calA_{t-1}]\right)=X_{T-1}.
\eaqn
Its differences are bounded: $|X_t-X_{t-1}|\leq B_t$. Hence,
it follows from Azuma's inequality that the probability that
$X_T$ exceeds some $\lambda>0$ is bounded by
$p=2\exp\big(-\tfrac{\lambda^2}{2\sum_t B_t^2}\big)$.
Requesting $\tfrac{\delta_T}{2}=p$ and solving for $\lambda$
gives the assertion.
\end{Proof}

The relation $\Expect \LOSS\foe \approxleq \Expect \LOSS\fpl$
follows immediately from the specification of the algorithm
$\FOE$.

\begin{Lemma} \label{lemma:foefpl}
$\big[\Expect \LOSS\foe \approxleq \Expect \LOSS\fpl\big]$ For
each $t\geq 1$, we have
$\Expect_t \loss_t\foe\leq(1-\gamma_t)\Expect_t\loss_t\fpl+\gamma_tB_t$.
\end{Lemma}

The next lemma relating $\Expect \LOSS\fpl$ and
$\Expect \LOSShat\fpl$ is technical but intuitively clear. It
states that in (conditional) expectation, the real loss
suffered by $\FPL$ is the same as the estimated loss. This is
simply because the loss estimate is unbiased. A combination
with the previous lemma was shown in \emcite{McMahan:04}.

\begin{Lemma} \label{lemma:fplfpl}
$\big[\Expect \LOSS\fpl \approxleq \Expect \LOSShat\fpl\big]$ For
each $t\geq 1$, we have
$\Expect_t \loss_t\fpl=\Expect_t\hat\loss_t\fpl$.
\end{Lemma}

Note that $\hat\loss_t\fpl$ is the loss $\hat\loss_t^I$
estimated by
$\FOE$, but for the expert $I=I_t\fpl$ chosen by $\FPL$.

\begin{Proof} Let $f^i_t=f^i_t(\calA_{t-1})=\Prob[I_t\fpl=i|\calA_{t-1}]$
be the probability distribution over actions $i$ which $\FPL$
uses at time $t$, depending on the past randomness
$\calA_{t-1}$. Let
$u_t=[1 \ldots 1]/n$ be the uniform distribution at time $t$ (for
non-uniform weights this will be replaced appropriately
later). Then
\baqn
\Expect_t[\hat\loss_t\fpl] =&
\gamma_t {\textstyle\sum_{i=1}^n}  f_t^i [(1-u_t^i)\cdot 0+u_t^i \hat\loss_t^i|_{r_t=1\wedge I\sfoe_t=i}]
\\=&
{\textstyle\sum_{i=1}^n}  f_t^i\loss_t^i =
\Expect_t[\loss_t\fpl],
\eaqn
where $\hat\loss_t^i|_{r_t=1\wedge
I\sfoe_t=i}=\loss_t^i/(u_t^i\gamma_t)$ is the estimated loss
under the condition that $\FOE$ decided to explore ($r_t=1$)
and chose action $I\foe_t=i$.
\end{Proof}

The following lemma from \emcite{Kalai:03} relates the losses
of $\FPL$ and $\IFPL$. We repeat the proof, since it is the
crucial and only step in the analysis where we have to be
careful with the upper loss bound $B_t$. Let
$\hat B_t=B_t (n/\gamma_t)$ denote the upper bound on the
instantaneous estimated losses.

\begin{Lemma} \label{lemma:fplifpl}
$\big[\Expect \LOSShat\fpl \approxleq \Expect \LOSShat\ifpl\big]$
$\Expect_t \hat\loss_t\fpl\leq\Expect_t\hat\loss_t\ifpl+
\gamma_t\eta_t \hat B_t^2$
holds for all $t\geq 1$.
\end{Lemma}

\begin{Proof}
If $r_t=0$, $\hat\loss_t= 0$ and thus
$\hat\loss_t\fpl=\hat\loss_t\ifpl$ holds. This happens with
probability $1-\gamma_t$. Otherwise we have
\beq
\label{eq:efpl}
\Expect_t\hat\loss_t\fpl=
\sum_{i=1}^n\int\eins_{I\sfpl_t=i}
\hat\loss_t^i d\mu(x),
\eeq
where $\mu$ denotes the (exponential) distribution of the
perturbations, i.e.\ $x_i:=q_t^i$ and density
$\mu(x):=\e^{-\|x\|_\infty}$. The idea is now that if action $i$
was selected by $\FPL$, it is -- because of the exponentially
distributed perturbation -- with high probability also selected by
$\IFPL$. Formally, we write $u^+=\max(u,0)$ for $u\in\RRR$,
abbreviate $\lambda=\hat \Loss\ltt+k/\eta_t$, and denote by
$\int\ldots d\mu(x_{\neq i})$ the integration leaving out the
$i$th action. Then, using
$\eta_t\lambda_i-x_i\leq\eta_t\lambda_j-x_j$ for all $j$ if
$I_t\fpl=i$ in the first line, and
$\hat B_t\geq \hat\loss_t^i-\hat\loss_t^j$ in the fourth line, we
get
\baqn
\label{eq:1}
\int&\eins_{I\sfpl_t=i}\hat\loss_t^i d\mu(x)
 =
  \int\int\limits_{x_i\geq\zwidths{\max\limits_{j\neq
  i}\{\eta_t(\lambda_i-\lambda_j)+x_j\}}}
  \hat\loss_t^i d\mu(x_i) d\mu(x_{\neq i})\\
& =  \int\hat\loss_t^i\;\e^{-(\max\limits_{j\neq i}\{\eta_t(\lambda_i-\lambda_j)+x_j\})^+}
   d\mu(x_{\neq i})\\
& \leq  \int\hat\loss_t^i\;\e^{\eta_t \hat B_t}\e^{-(\max\limits_{j\neq i}\{\eta_t(\lambda_i-\lambda_j)+x_j\}+\eta_t \hat B_t)^+}
   d\mu(x_{\neq i})\\
& \leq\e^{\eta_t \hat B_t} \int\hat\loss_t^i\;\e^{-(\max\limits_{j\neq i}\{\eta_t(\lambda_i+\hat\loss_t^i-\lambda_j-\hat\loss_t^j)+x_j\})^+}
   d\mu(x_{\neq i})\\
& =\e^{\eta_t \hat B_t}\int\eins_{I\sifpl_t=i}
\hat\loss_t^i d\mu(x).
\eaqn
Summing over $i$ and using the analogue of (\ref{eq:efpl})
for $\IFPL$, we see that if $r_t=1$, then
$\Expect_t\hat\loss_t\fpl\leq
\e^{\eta_t \hat B_t}\Expect_t\hat\loss_t\ifpl$ holds. Thus
$\Expect_t\hat\loss_t\ifpl\geq
\e^{-\eta_t \hat B_t}\Expect_t\hat\loss_t\fpl
\geq(1-\eta_t \hat B_t)\Expect_t\hat\loss_t\fpl
\geq\Expect_t\hat\loss_t\fpl-\eta_t\hat B_t^2$.
The assertion now follows by taking expectations w.r.t\ $r_t$.
\end{Proof}

The next lemma relates the losses of $\IFPL$ and the best
action in hindsight. For an oblivious adversary (which means
that the adversary's decisions do not depend on our past
actions), the proof was given in \emcite{Kalai:03}. An
additional step is necessary for an adaptive adversary. We
omit the proof here, the reader may reconstruct it from the
proof of Lemma \ref{lemma:ifplbehtau}.

\begin{Lemma} \label{lemma:ifplbeh}
$\big[\Expect \LOSShat\ifpl \approxleq \LOSShat^{\xwidehat\best}\big]$ Assume
decreasing learning rate $\eta_t$ and $\sum_i\e^{-k^i}\leq 1$. For
all $T\geq 1$ and $1\leq i\leq n$, we have $\sum_{t=1}^T\Expect_t
\hat \loss\ifpl_t\leq \hat\loss\leqT^i+\frac{k^i}{\eta_T}$
(recall that $\hat\loss\leqT^i$ is a random variable depending
on $\calA_t$).
\end{Lemma}

Finally, we give a relation between the estimated and true
losses, adapted from \emcite{McMahan:04}.

\begin{Lemma} \label{lemma:behbeh}
$\big[\LOSShat^{\xwidehat\best} \approxleq \LOSS^{\best}\big]$
For each $T\geq 1$, $\delta_T\in(0,1)$, and $1\leq i\leq n$,
w.p.\ at least $1-\tfrac{\delta_T}{2}$ we have
\beq
\label{eq:behbeh}
\hat \Loss\leqT^i
\leq \Loss\leqT^i+
\tsqrt{(2\ln\tfrac{4}{\delta_T}){\textstyle\sum\nolimits_{t=1}^{T}
\hat B_t^2}}.
\eeq
\end{Lemma}

\begin{Proof}
$X_t=\hat \Loss\leqt^i-\Loss\leqt^i$ is a martingale, since
\baqn
\Expect[X_t|\calA_{t-1}]&=\Expect[\hat \Loss^i\leqt|\calA_{t-1}]-\Loss\leqt^i
\\&=
X_{t-1}+\Expect[\hat\loss^i_t|\calA_{t-1}]-\loss_t^i=X_{t-1}.
\eaqn
Its differences are bounded:
$|X_t-X_{t-1}|\leq\hat B_t$. By Azuma's inequality,
its actual value at time $T$ does not exceed
$\tsqrt{(2\ln\tfrac{4}{\delta_T}){\textstyle\sum\nolimits_{t=1}^{T}
\hat B_t^2}}$ w.p.\ $1-\tfrac{\delta_T}{2}$.
\end{Proof}

We now combine the above results and derive an upper bound
on the expected regret of $\FOE$ against an adaptive
adversary.

\begin{Theorem} \label{th:uniform} {\upshape[$\FOE$ against
an adaptive adversary]} Let $n$ be finite and $k^i=\ln n$ for
all $1\leq i\leq n$. Let $\eta_t$ be decreasing, and
$\loss_t\in[0,B_t]^n$ some possibly adaptive assignment of loss
vectors. Then for all experts $i$,
\baqn
&\Loss\foe\leqT\leq
\loss\leqT^i+\sqrt{(2\ln\tfrac{4}{\delta_T})}
\left(\sqrt{\sum_{t=1}^{T}
\tfrac{B_t^2n^2}{\gamma_t^2}}
+\sqrt{{\sum_{t=1}^{T} B_t^2}}\right)\\
&\ +\tfrac{\ln n}{\eta_T}+\sum_{t=1}^T\tfrac{\eta_t B_t^2
n^2}{\gamma_t} + \sum_{t=1}^T\gamma_t B_t
\ \wprob 1-\delta_T \und\\
&\Expect \Loss\foe\leqT\leq
\loss\leqT^i+\tfrac{\ln n}{\eta_T}+\sum_{t=1}^T\tfrac{\eta_t B_t^2 n^2}{\gamma_t}
+ \sum_{t=1}^T\gamma_t B_t\\
&\ +\sqrt{(2\ln\tfrac{4}{\delta_T}){\sum_{t=1}^{T}
\tfrac{B_t^2n^2}{\gamma_t^2}}}+
\tfrac{\delta_T}{2} {\sum_{t=1}^{T}\tfrac{B_t n}{\gamma_t}}
.
\eaqn
\end{Theorem}

\begin{Proof}
The first high probability bound follows by summing up all
excess terms in the above lemmas, observing that $\hat B_t=B_t
(n/\gamma_t)$. For the second bound on the expectation, we
take expectations in Lemmas
\ref{lemma:foefpl}-\ref{lemma:ifplbeh}, while Lemma
\ref{lemma:foefoe} is not used. For Lemma \ref{lemma:behbeh},
a statement in expectation is obtained as follows:
(\ref{eq:behbeh}) fails w.p.\ at most $\frac{\delta_T}{2}$, in
which case
$\hat \Loss^i\leqT-\Loss^i\leqT\leq\sum\nolimits_{t=1}^{T}\hat B_t$.
\end{Proof}

\begin{Cor} \label{cor:uniform}
Under the conditions of Theorem \ref{th:uniform},
\bqan
(i) & B_t\equiv 1 &\Rightarrow\
\Expect \Loss\foe\leqT\leq
\loss\leqT^i+O(n^2T^{\frac{3}{4}}\sqrt{\ln T}),\\
(ii) & B_t\equiv 1 &\Rightarrow\ \Loss\foe\leqT\leq
\loss\leqT^i+O(n^2T^{\frac{3}{4}}\sqrt{\ln T}), \\
(iii) & B_t=t^{\frac{1}{8}} &\Rightarrow\
\Expect \Loss\foe\leqT\leq
\loss\leqT^i+O(n^2T^{\frac{7}{8}}\sqrt{\ln T}), \\
(iv) & B_t=t^{\frac{1}{8}} &\Rightarrow\ \Loss\foe\leqT\leq
\loss\leqT^i+O(n^2T^{\frac{7}{8}}\sqrt{\ln T}),
\eqan
for all $i$ and $T$. Here, $(ii)$ and $(iv)$ hold with
probability $1-T^{-2}$. Moreover, in both cases (bounded and
growing $B_t$) $\FOE$ is asymptotically optimal, i.e.\
\beqn
\limsup_{T\to\infty} \tfrac{1}{T}\big(
\Loss\foe\leqT-\min_i\loss^i\leqT\big)\leq 0
\quad\mbox{ almost surely.}
\eeqn
\end{Cor}

$B_t=t^{\frac{1}{8}}$ in $(iii)$ and $(iv)$ is
just one choice to achieve asymptotic optimality while the
losses may grow unboundedly. Asymptotic optimality is
sometimes termed \emph{Hannan-consistency}, in particular if
the limit equals zero. We only show the upper bound.

\begin{Proof}
$(i)$ and $(ii)$ follow by applying the previous theorem to
$\eta_t=t^{-\frac{1}{2}}$,
$\gamma_t=t^{-\frac{1}{4}}$, $\delta_T=T^{-2}$,
and observing $\sum_{t=1}^T t^\alpha\leq\int_0^{T+1}
t^\alpha\leq 2(T+1)^{1+\alpha}$ for
$\alpha\geq -\frac{1}{2}$. In order to obtain $(iii)$ and
$(iv)$, set $\eta_t=t^{-\frac{3}{4}}$,
$\gamma_t=t^{-\frac{1}{4}}$, and $\delta_T=T^{-2}$. The
asymptotic optimality finally follows from the Borel-Cantelli
Lemma, since
\beqn
\textstyle
\Prob\left[
\frac{1}{T}(\Loss\foe\leqT-\min_i\loss^i\leqT)>
C T^{-\frac{1}{8}}\sqrt{\ln T}\,\right]\leq\frac{1}{T^2}
\eeqn
for an appropriate $C>0$ according to $(ii)$ and $(iv)$.
\end{Proof}

\section{Infinitely Many Experts and Arbitrary Priors}
\label{sec:arbitrary}

\begin{figure}[t!]
\begin{center}
\fbox{
\begin{minipage}{\algowidth}
For $t=1,2,3,\ldots$\\
Sample $r_t\in\{0,1\}$ independently s.t. $P[r_t=1]=\gamma_t$\\
If $r_t=0$ Then\\
\ii Invoke $\FPL{}^\tau(t)$ and play its decision\\
\ii Set $\hat\loss_t^i=0$ for $i\in\{t\geq\tau\}$\\
Else\\
\ii Sample $I_t$ w.r.t.\ $u_t$ in (\ref{eq:unonu}) and play $I:=I_t\foetau$\\
\ii Set $\hat\loss_t^I=\loss_t^I/(u_t^I \gamma_t)$ and $\hat\loss_t^i=0$ for $i\in\{t\geq\tau\}\setminus\{I\}$ \\
Set $\hat\loss_t^i=\hat B_t$ for $i\not\in\{t\geq\tau\}$
\end{minipage}}
\caption{The algorithm $\FOE{}^\tau$}
\label{fig:foetau}
\end{center}
\end{figure}

\begin{figure}[t!]
\begin{center}
\fbox{
\begin{minipage}{\algowidth}
Sample $q_t^i\stackrel{d.}{\sim}\mbox{\textit{Exp}}$
independently for $i\in\{t\geq\tau\}$\\
select and play
$I_t\fpl=\arg\min\limits_{i:t\geq\tau}
\{\eta_t\hat \Loss^i\ltt+k^i-q_t^i\}$
\end{minipage}}
\caption{The algorithm $\FPL{}^\tau(t)$}
\label{fig:fpltau}
\end{center}
\end{figure}

The following considerations are valid for both finitely and
infinitely many experts with arbitrary prior weights $w^i$.
For notational convenience, we write $n=\infty$ in the latter
case. When admitting infinitely many experts, two difficulties arise:
Since the prior
weights of the experts sum up to one and thus become
arbitrarily small, the estimated losses -- obtained by
dividing by these weights -- would possibly get arbitrarily
large. We therefore introduce, for each expert $i$, a time
$\tau^i\geq 1$ at which the expert enters the game. All
algorithms $\FOE$,
$\FPL$, $\IFPL$ are substituted by counterparts
$\FOE{}^\tau$, $\FPL{}^\tau$, $\IFPL{}^\tau$
which use expert $i$ only for $t\geq\tau^i$. Thus, the maximum
estimated loss possibly assigned to these \emph{active}
experts is
\beq
\label{eq:fplifpltau}
\hat B_t={B_t}/[{\gamma_t\min\{w^i:t\geq\tau^i\}}].
\eeq
We denote the set of active experts at time $t$ by
$\{t\geq\tau\}=\{i:t\geq\tau^i\}$. Experts which have not yet
entered the game are given an estimated loss of $\hat B_t$.
This also solves the computability problem: Since at every
time $t$ only a finite number of experts is involved,
$\FOE{}^\tau$ is computable (if each expert is). The
algorithms $\FOE{}^\tau$ and $\FPL{}^\tau$ are specified
in Figures \ref{fig:foetau} and \ref{fig:fpltau}.

Again, the analysis follows the outline (\ref{eq:diagram}).
Lemmas \ref{lemma:foefoe}--\ref{lemma:fplifpl} have equivalent
counterparts, the proofs of which remain almost unchanged. In
Lemma \ref{lemma:fplfpl}, the ``uniform" distribution over
experts
$u_t$ now becomes
\beq
\label{eq:unonu}
u_t^i={w^i\eins_{t\geq\tau^i}/[\textstyle \sum_j
w^j\eins_{t\geq\tau^j}]}.
\eeq
The upper bound on the estimated loss
$\hat B_t$ in Lemma \ref{lemma:fplifpl} is given by
(\ref{eq:fplifpltau}). We only need to prove assertions
corresponding to Lemmas \ref{lemma:ifplbeh} and
\ref{lemma:behbeh}.

\begin{Lemma} \label{lemma:ifplbehtau}
$\big[\Expect \LOSShat\ifpltau \approxleq \LOSShat^{\xwidehat\best{}^\tau}\big]$
Assume that $\sum_i\e^{-k^i}\leq 1$ and $\tau^i$ depends
monotonically on $k^i$, i.e.\
$\tau^i\geq\tau^j$ if and only if $k^i\geq k^j$. Assume
decreasing learning rate $\eta_t$.
For all $T\geq 1$ and all $1\leq i\leq n$, we have
\beqn
\sum_{t=1}^T\Expect_t \hat \loss\ifpltau_t\leq
\hat\loss\leqT^i+{\textstyle\frac{k^i+1}{\eta_T}}.
\eeqn
\end{Lemma}

\begin{Proof}
This is a modification of the corresponding proofs in
\emcite{Kalai:03} and \emcite{Hutter:04expert}. We may fix the
randomization $\calA$ and suppress it in the notation. Then we
only need to show
\beq
\label{eq:showtau}
\Expect \hat
\Loss\ifpltau\leqT\leq\min\limits_{1\leq i\leq n}
\{\hat\loss\leqT^i+{\textstyle\frac{k^i+1}{\eta_T}}\},
\eeq
where the expectation is with respect to $\IFPL$'s randomness $q_{1:T}$.

Assume first that the adversary is oblivious. We define an
algorithm $A$ as a variant of $\IFPL{}^\tau$ which samples
only one perturbation vector $q$ in the beginning and uses
this in each time step, i.e.\ $q_t\equiv q$. Since the
adversary is oblivious, $A$ is equivalent to
$\IFPL{}^\tau$ in terms of expected performance. This is all
we need to show (\ref{eq:showtau}). Let
$\eta_0=\infty$ and
$\lambda_t=\hat\loss_t+(k-q)
\big(\frac{1}{\eta_t}-\frac{1}{\eta_{t-1}}\big)$, then
$\lambda\leqt=\hat\loss\leqt+\frac{k-q}{\eta_t}$.
Recall $\{t\geq\tau\}=\{i:t\geq\tau^i\}$. We argue by
induction that for all
$T\geq 1$,
\beq
\label{eq:ifplbehtau}
\sum_{t=1}^T \lambda_t^A\leq
\min_{T\geq\tau}\lambda\leqT^i+
\max_{T\geq\tau}\big\{{\tfrac{q^i-k^i}{\eta_T}}\big\}.
\eeq
This clearly holds for $T=0$. For the induction step, we have
to show
\baq
\label{eq:showtau2}
&\min_{T\geq\tau}\lambda\leqT^i+
\max_{T\geq\tau}{\tfrac{q^i-k^i}{\eta_T}}+
\lambda_{T+1}^A
\leq
\lambda\leqT^{I^A_{T+1}}
\\ \nonumber
&+\max_{T+1\geq\tau}{\tfrac{q^i-k^i}{\eta_{T+1}}}
+\lambda_{T+1}^{I^A_{T+1}}
 = \min_{T+1\geq\tau}\lambda_{1:T+1}^i+
\max_{T+1\geq\tau}{\tfrac{q^i-k^i}{\eta_{T+1}}}.
\eaq
The inequality is obvious if $I_{T+1}^A\in\{T\geq\tau\}$.
Otherwise, let
$J=\arg\max\big\{q^i-k^i:i\in\{T\geq\tau\}\big\}$. Then
\baqn
\min_{T\geq\tau}&\lambda\leqT^i+
  \max_{T\geq\tau}\big\{{\tfrac{q^i-k^i}{\eta_T}}\big\}
\leq
  \lambda\leqT^J+\tfrac{q^J-k^J}{\eta_T}
  = \sum_{t=1}^T\hat\loss_t^J
  \\&\leq \sum_{t=1}^T\hat B_t
 =
  \sum_{t=1}^T\hat\loss_t^{I_{T+1}^A}
  \leq \lambda\leqT^{I^A_{T+1}}+
  \max_{T+1\geq\tau}\big\{{\tfrac{q^i-k^i}{\eta_{T+1}}}\big\}
\eaqn
shows (\ref{eq:showtau2}). Rearranging terms in
(\ref{eq:ifplbehtau}), we see
\beqn
  \sum_{t=1}^T \hat\loss_t^A\leq
  \min_{T\geq\tau}\lambda\leqT^i \!+\!
  \max_{T\geq\tau^i}\big\{{\textstyle\frac{q^i-k^i}{\eta_T}}\big\}
  \!+\!\! \sum_{t=1}^T
(q\!-\!k)^{I_t^A}\big({\textstyle\frac{1}{\eta_t}\!-\!\frac{1}{\eta_{t-1}}}\big).
\eeqn
The assertion (\ref{eq:showtau}) -- still for oblivious
adversary and $q_t\equiv q$ -- then follows by taking
expectations and using
\baqn
&\Expect\min_{T\geq\tau}\lambda\leqT^i
\!\leq  \min_{T\geq\tau}\{\hat\loss\leqT^i\!+\!\tfrac{k^i}{\eta_T}
\!-\!\Expect {\textstyle\frac{q^i}{\eta_T}} \}
\!\!\stackrel{(*)}{\leq} \min_{\zwidths{1\leq i\leq n}}
\{\hat\loss\leqT^i\!+\!{\textstyle\frac{k^i-1}{\eta_T}}\}\\
&\mbox{and }\Expect\sum_{t=1}^T
(q-k)^{I_t^A}\big(\tfrac{1}{\eta_t}-\tfrac{1}{\eta_{t-1}}\big)
\!\leq\!
\Expect\max_{T\geq\tau}\big\{{\textstyle\frac{q^i-k^i}{\eta_T}}\big\}
\!\leq\! {\textstyle\frac{1}{\eta_T}}.
\eaqn
Here, $(*)$ holds because
$\tau^i$ depends monotonically on $k^i$, and $\Expect q^i=1$, and
maximality of $\hat\loss_{1:T}^i$ for $T<\tau_i$. The last
inequality can be proven by an application of the union bound
\cite[Lem.1]{Hutter:04expert}.

Sampling the perturbations $q_t$ independently is equivalent
under expectation to sampling $q$ only once. So assume that
$q_t$ are sampled independently, i.e.\ that
$\IFPL{}^\tau$ is played against an oblivious
adversary: (\ref{eq:showtau}) remains valid. In the last step,
we argue that then (\ref{eq:showtau}) also holds for an
\emph{adaptive} adversary. This is true because the future
actions of $\IFPL{}^\tau$ do not depend on its past actions,
and therefore the adversary cannot gain from deciding after
having seen $\IFPL{}^\tau$'s decisions. (For details see
\npcite{Hutter:04expertx}. Note the subtlety that the future
actions of $\FOE{}^\tau$ would depend on its past actions.)
\end{Proof}

\begin{Lemma} \label{lemma:behbehtau}
$\big[\LOSShat^{\xwidehat\best{}^\tau} \approxleq \LOSS^{\best}\big]$
For each $T\geq 1$, $\delta_T\in(0,1)$, and $1\leq i\leq n$,
we have
$\hat \Loss\leqT^i
\leq \Loss\leqT^i + \tsqrt{(2\ln\tfrac{4}{\delta_T})\sum_{t=1}^{T}
\hat B_t^2} +\sum_{t=1}^{\tau^i-1}\hat B_t$ w.p.
$1-\tfrac{\delta_T}{2}$.
\end{Lemma}

This corresponds to Lemma \ref{lemma:behbeh}. The proof
proceeds in a similar way: we have to note that
$\hat\Loss\leqt^i-\Loss\leqt^i$ is a martingale only for
$t\geq\tau^i$, and $\hat \Loss_{<\tau^i}^i$ exceeds $\Loss_{<\tau^i}^i$
by at most $\sum_{t=1}^{\tau^i-1}\hat B_t$. Then the following
theorem corresponds to Theorem \ref{th:uniform} and is proven
likewise.

\begin{Theorem} \label{th:general} {\upshape[$\FOE{}^\tau$ against
an adaptive adversary]} Let $n$ be finite or infinite,
$\sum_i\e^{-k^i}\leq 1$, $\tau^i$ depend monotonically on
$k^i$, and the learning rate $\eta_t$ be decreasing.
Let $\loss_t$ some possibly adaptive assignment of (true) loss
vectors satisfying
$\|\loss_t\|_\infty\leq B_t$. Then for all experts $i$, we
have
\baqn
\Loss\foetau\leqT \!\!\leq\,&
\loss\leqT^i
+\sqrt{(2\ln\tfrac{4}{\delta_T})}\left(\sqrt{{\sum_{t=1}^{T}
\tfrac{B_t^2}{\gamma_t^2(w^{*}_t)^2}}}
+\sqrt{{\sum_{t=1}^{T} B_t^2}}\right)
\\&
+\!\tfrac{k^i+1}{\eta_T}+
\sum_{t=1}^{\tau^i-1} \tfrac{B_t}{\gamma_t w^{*}_t} +\!
\sum_{t=1}^T\tfrac{\eta_t B_t^2}{\gamma_t (w^{*}_t)^2}
+\!\sum_{t=1}^T\gamma_t B_t
\eaqn
with probability $1-\delta_T$, where
$w^{*}_t=\min\{w^i:t\geq\tau^i\}$. A corresponding statement
holds for the expectation (compare Theorem \ref{th:uniform}).
\end{Theorem}

\begin{Cor} \label{cor:arbitrary}
Assume the conditions of Theorem \ref{th:general}. Then for
all $i$ and $T$, the following holds w.p. $1-\delta_T$.
\baqn
(i) \ & B_t\equiv 1,\tau^i=\lceil(w^i)^{-8}\rceil \\
& \Rightarrow
\Loss\foe\leqT\leq \loss\leqT^i+O\big((\tfrac{1}{w^i})^{11} + T^{\frac{7}{8}}\sqrt{\ln T}\big),
\und\\
(ii) \ & B_t=t^{\frac{1}{16}},\tau^i=\lceil(w^i)^{-16}\rceil \\
& \Rightarrow
\Loss\foe\leqT\leq \loss\leqT^i+O\big((\tfrac{1}{w^i})^{22} +
T^{\frac{7}{8}}\sqrt{\ln T}\big).
\eaqn
Corresponding assertions are true for the expectation (compare
Corollary \ref{cor:uniform}). In both cases (bounded and
growing
$B_t$) $\FOE$ is asymptotically optimal w.r.t.\
each expert:
$
\limsup_{T\to\infty} \tfrac{1}{T}\big(\Loss\foe\leqT-\loss^i\leqT\big)\leq 0
$
a.s.\ for all $i$.
\end{Cor}

\begin{Proof}
Let $\eta_t=t^{-\frac{3}{4}}$,
$\gamma_t=t^{-\frac{1}{4}}$, and $\delta_T=T^{-2}$. For
$\tau^i=\lceil(w^i)^{-\alpha}\rceil$ and
$B_t=t^\beta$, we have
$w^{*}_T=\min\{w^i:T\geq\lceil(w^i)^{-\alpha}\rceil\}\geq
\min\{w^i:T^{-\frac{1}{\alpha}}\leq w^i\rceil\}\geq
T^{-\frac{1}{\alpha}}$ and
\beqn
\sum_{t=1}^{\tau^i-1}\hat B_t \leq
(\tau^i-1)\hat B_{\tau^i-1} \leq
\tfrac{(w^i)^{-\alpha} B_{\tau^i-1}}{\gamma_{\tau^i-1}w^{*}_{\tau^i-1}}
\leq \tfrac{(w^i)^{-\alpha}
(w^i)^{-\alpha\beta}}{(w^i)^{\frac{\alpha}{4}}w^i}
\eeqn
(observe $w^{*}_{\tau^i-1}\geq
(\tau^i-1)^{-\frac{1}{\alpha}}\geq
(w^i)^{(-\alpha)(-\frac{1}{\alpha})}$). Then set
$\alpha=8$, $\beta=0$, for $(i)$ and
$\alpha=16$, $\beta=\frac{1}{16}$ for $(ii)$. Asymptotic
optimality is shown as in Corollary \ref{cor:uniform}.
\end{Proof}

\section{Active Expert Problems and a Universal Master Algorithm}
\label{sec:active}

If the adversary's goal is just to maximize our (expected)
regret, then it is well known what he can achieve (at least
for uniform prior, see e.g.\ the lower bound in
\npcite{Cesa:97,Auer:02bandit}). We are interested in different
situations. An example is the repeated playing of the
``Prisoner's dilemma" against the Tit-for-Tat\footnote{In the
prisoner's dilemma, two players both decide independently if
thy are \emph{cooperating (C)} or \emph{defecting (D)}. If
both play \emph{C}, they get both a small loss, if both play
\emph{D}, they get a large loss. However, if one plays \emph{C}
and one \emph{D}, the cooperating player gets a very large
loss and the defecting player no loss at all. Thus defecting
is a \emph{dominant} strategy. A Tit-for-Tat player play
\emph{C} in the first move and afterwards the opponent's
respective preceding move.} strategy \cite{Farias:03}. If we
use two strategies as experts, namely ``always cooperate" and
``always defect", then it is clear that always cooperating
will have the better long-term reward. It is also clear that a
standard expert advice or bandit master algorithm will not
discover this, since it compares only the losses in one step,
which are always lower for the defecting expert.

We therefore propose to give the control to a selected expert
for \emph{periods of increasing length}. Precisely, we
introduce a new time scale $\tilde t$ at which we have single
games with losses $\tilde\loss_{\tilde t}$. The master's time
scale $t$ does not coincide with $\tilde t$. Instead, at each
$t$, the master gives control to the selected expert $i$ for
$\tilde T_t$ single games and receives loss
$\loss_t^i=\sum_{\tilde t=\tilde t(t)}^{\tilde t(t)+\tilde T_t-1}
\tilde\loss_{\tilde t}^i$. Assume that the game has bounded
instantaneous losses $\tilde\loss_{\tilde t}^i\in[0,1]$. Then
the master algorithm's instantaneous losses are bounded by
$\tilde T_t$. We denote this algorithm by $\FOE_{\smash{\tilde T}}$
or $\FOE{}_{\smash{\tilde T}}^\tau$.

\begin{Cor}
Assume $\FOE_{\smash{\tilde T}}$ (or $\FOE{}_{\smash{\tilde
T}}^\tau$, respectively) plays a repeated game with bounded
instantaneous losses $\tilde\loss_{\tilde t}^i\in[0,1]$. Let
the exploration and learning rates be
$\gamma_t=t^{-\frac{1}{4}}$ and
$\eta_t=t^{-\frac{3}{4}}$. In case of uniform prior, choose
$\tilde T_t=\lfloor t^{\frac{1}{8}}\rfloor$ ($\tau^i\equiv 0$).
In case of arbitrary prior let
$\tilde T_t=\lfloor t^{\frac{1}{16}}\rfloor$ and
$\tau^i=\lceil(w^i)^{-16}\rceil$.
Then for all experts $i$ and all
$\tilde T$, suppressing the dependence on the prior of expert $i$, we
have
\bqan
  \Loss\foett_{1:\smash{\tilde T}} &\leq& \Loss_{1:\tilde T}^i+
  O(\tilde T^{\frac{9}{10}})
\wprob 1-\tilde T^2
\und\\
  \Expect \Loss\foett_{1:\smash{\tilde T}} &\leq& \Loss_{1:\tilde T}^i+O(\tilde
  T^{\frac{9}{10}}).
\eqan
Consequently,
$\limsup_{T\to\infty} (\Loss\foett_{1:\smash{\tilde T}}-\loss^i_{1:\smash{\tilde T}})/\tilde
T\leq0$ almost surely. The rate of convergence is at least
$\tilde T^{-\frac{1}{10}}$. The same assertions hold for $\FOE{}_{\smash{\tilde T}}^\tau$.
\end{Cor}

\begin{Proof}
This follows from changing the time scale from $t$ to
$\tilde t$ in Corollaries \ref{cor:uniform} and
\ref{cor:arbitrary}: $\tilde t$ is of order
$t^{1+\frac{1}{8}}$ in the uniform case and
$t^{1+\frac{1}{16}}$ in the general case. Then the
bounds are $\tilde T^{\frac{8}{9}}\sqrt{\ln \tilde T}$ in the
former and $\tilde T^{\frac{15}{17}}\sqrt{\ln
\tilde T}$ in the latter case. Both are upper
bounded by $\tilde T^{\frac{9}{10}}$.
\end{Proof}

Broadly spoken, this means that $\FOE_{\smash{\tilde T}}$
performs asymptotically as well as the best expert. Asymptotic
guarantees for the Strategic Experts Algorithm have been
derived by \aunpcite{Farias:03}. Our results approve upon this
by providing a rate of convergence. One can give further
corollaries, e.g.\ in terms of flexibility as defined by
\aunpcite{Farias:03}.

It is also possible to specify a \emph{universal} experts
algorithm. To this aim, let expert $i$ be derived from the
$i$th program $p^i$ of some fixed universal Turing machine. The
$i$th program can be well-defined, e.g.\ by representing
programs as binary strings and lexicographically ordering them
\cite{Hutter:04uaibook}. Before the expert is consulted, the
relevant input is written to the input tape of the
corresponding program. If the program halts, the appropriate
number of first bits is interpreted as the expert's
recommendation. E.g.\ if the decision is binary, then the
first bit suffices. (If the program does not halt, we may for
well-definedness just fill its output tape with zeros.)  Each
expert is assigned a prior weight by
$w^i=2^{-\mbox{\scriptsize{length}}(p^i)}$, where
$\mbox{length}(p^i)$ is the length of the
corresponding program and we assume the program tape to be
binary. This construction parallels the definition of
Solomonoff's \emph{universal prior} \yrcite{Solomonoff:78}.
This has been used to define a universal agent AIXI in a quite
different way by \emcite{Hutter:04uaibook}. Note that like the
universal prior and AIXI, our universal agent is not
computable, since we cannot check if a program halts. It is
however straightforward to impose a bound on the computation
time which for instance increases rapidly in $t$. If used with
computable experts, the algorithm is computationally feasible.
The universal master algorithm performs well with respect to
\emph{any} computable strategy.

\begin{Cor}
Assume the universal set of experts specified in the last
paragraph. If $\FOE{}_{\smash{\tilde T}}^\tau$ is applied with
$\gamma_t=t^{-\frac{1}{4}}$, $\eta_t=t^{-\frac{3}{4}}$,
$\tilde T_t=\lfloor t^{\frac{1}{16}}\rfloor$, and
$\tau^i=\lceil(w^i)^{-16}\rceil$, then it performs
asymptotically at least as good as \emph{any computable
expert} $i$. The rate of convergence is exponential in the
complexity
$k^i$ and proportional to $\tilde T^{-\frac{1}{10}}$.
\end{Cor}

\section{Discussion}
\label{sec:discussion}

For large or infinite expert classes, the bounds we have
proven are irrelevant in practice, although asserting almost
sure optimality and even a convergence rate: the exponential
of the complexity is far too huge. Imagine for instance a
moderately complex task and some good strategy, which can be
coded with mere 500 bits. Then its weight is $2^{-500}$, a
constant which is not distinguishable from zero in all
practical situations. Thus, it seems that the bounds can be
relevant at most for small expert classes with uniform prior.
This is a general shortcoming of bandit experts algorithms:
For uniform prior a lower bound on the expected loss which is
linear in $\sqrt n$ has been proven \cite{Auer:02bandit}.

If the bounds are not practically relevant, maybe the
algorithms are so? We leave this interesting question
unanswered. Intuitively, it might seem that the algorithms
proposed here are too much tailored towards worst-case bounds
and fully adversarial setups. For example, the exploration
rate of $t^{-\frac{1}{4}}$ is quite high. Master algorithms
which are less ``cautious" might perform better for many
practical problems. Finally, it would be nice to investigate
the differences between the proposed expert style approach and
other definitions of universal agents, such as by
\emcite{Hutter:04uaibook}.

\textbf{Acknowledgement}: This work was supported by SNF grant 2100-67712.02.


\begin{small}

\end{small}

\end{document}